\documentclass{article}

\usepackage{arxiv}

\usepackage[utf8]{inputenc} % allow utf-8 input
\usepackage[T1]{fontenc}    % use 8-bit T1 fonts
\usepackage{url}            % simple URL typesetting
\usepackage{booktabs}       % professional-quality tables
\usepackage{amsfonts}       % blackboard math symbols
\usepackage{nicefrac}       % compact symbols for 1/2, etc.
\usepackage{microtype}      % microtypography
\usepackage{subcaption}
\usepackage{graphicx}
\usepackage{xcolor}   
\usepackage{color}
\usepackage{wrapfig}

\definecolor{cvprblue}{rgb}{0.21,0.49,0.74}
\usepackage[pagebackref,breaklinks,colorlinks,citecolor=cvprblue]{hyperref}
\usepackage{doi}

\usepackage{amsmath}
\usepackage{amssymb}
\usepackage{multirow}
\usepackage{multicol}
\usepackage{comment}

\usepackage{algorithm}
\usepackage{algpseudocode}

\definecolor{ered}{rgb}{0.72, 0.16, 0.2}

\newcommand{\q}[1]{``#1''} % Fancy quotes 
\newcommand{\std}[1]{\textbf{\scriptsize\textcolor{gray}{$\pm#1$}}} % STD +- in tables

\definecolor{best}{RGB}{168, 226, 219}
\definecolor{drop}{RGB}{205, 95, 95}
\definecolor{gray}{RGB}{150, 150, 150}

\usepackage{cleveref}       % smart cross-referencing

\newcommand*\samethanks[1][\value{footnote}]{\footnotemark[#1]}

\begin{document}

\title{Distilling Datasets Into Less Than One Image}
\author{Asaf Shul\thanks{Equal contribution} \qquad Eliahu Horwitz\samethanks \qquad Yedid Hoshen\\
  School of Computer Science and Engineering\\
  The Hebrew University of Jerusalem, Israel\\
  \url{https://vision.huji.ac.il/podd/}\\
  \texttt{\{asaf.shul, eliahu.horwitz, yedid.hoshen\}@mail.huji.ac.il} \\
}

\maketitle

\begin{figure*}[h]
\centering
\includegraphics[width=0.99\linewidth]{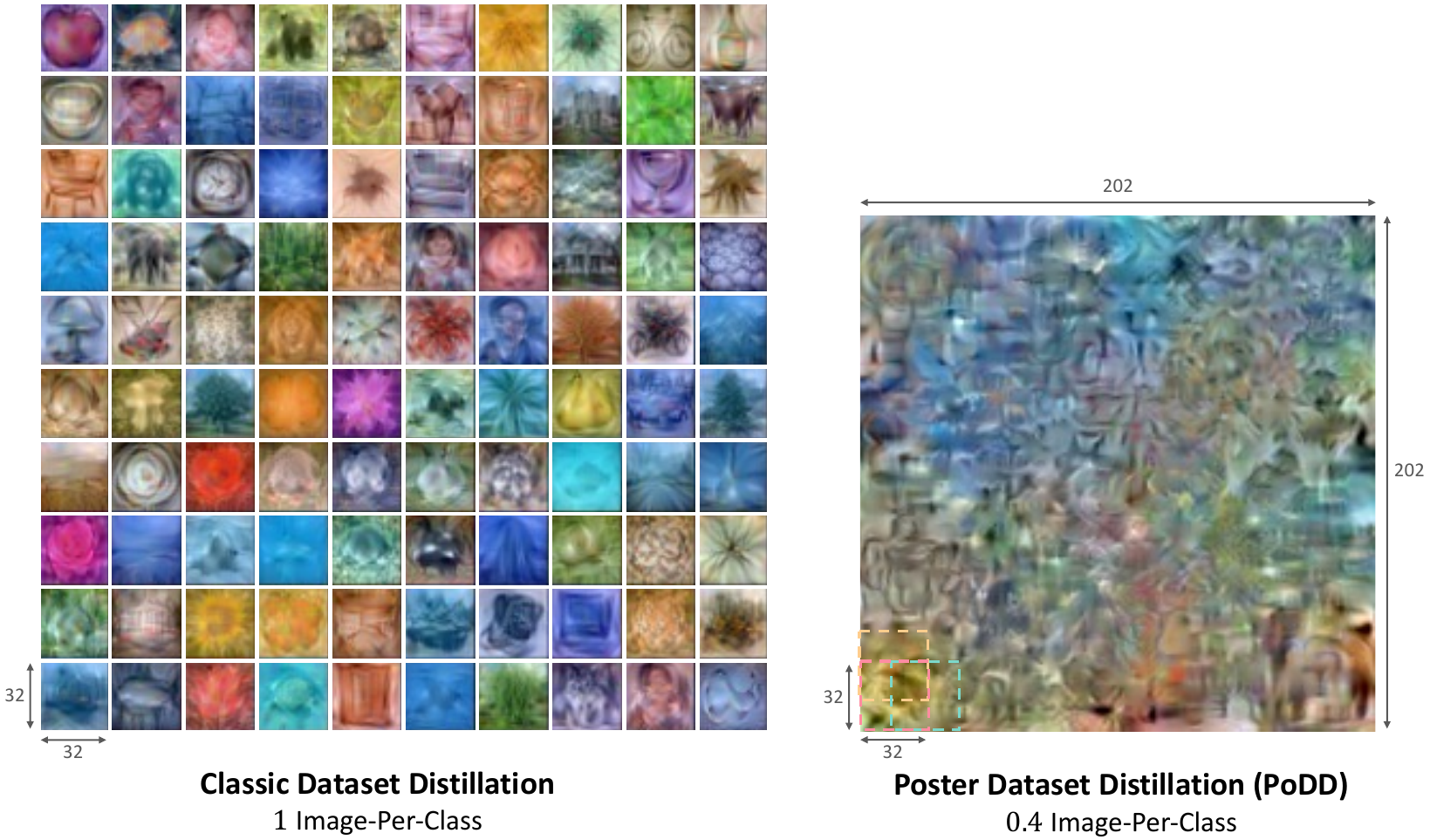}
\caption{\textit{\textbf{Poster Dataset Distillation (PoDD):}} We propose PoDD, a new dataset distillation setting for a tiny, under $1$ image-per-class (IPC) budget. In this example, the standard method attains an accuracy of $35.5\%$ on CIFAR-100 with approximately $100k$ pixels, PoDD achieves an accuracy of $35.7\%$ with less than half the pixels (roughly $40k$)
}
\label{fig:1_vs_04_ipc}
\end{figure*}

\begin{abstract}
Dataset distillation aims to compress a dataset into a much smaller one so that a model trained on the distilled dataset achieves high accuracy. Current methods frame this as maximizing the distilled classification accuracy for a budget of $K$ distilled \textit{images-per-class}, where $K$ is a positive integer. In this paper, we push the boundaries of dataset distillation, compressing the dataset into less than an image-per-class. It is important to realize that the meaningful quantity is not the number of distilled images-per-class but the number of distilled pixels-per-dataset. We therefore, propose \textit{Poster Dataset Distillation} (PoDD), a new approach that distills the entire original dataset into a single poster. The poster approach motivates new technical solutions for creating training images and learnable labels. Our method can achieve comparable or better performance with less than an image-per-class compared to existing methods that use one image-per-class. Specifically, our method establishes a new state-of-the-art performance on CIFAR-10, CIFAR-100, and CUB200 using as little as $0.3$ images-per-class. 

\end{abstract}

\section{Introduction}
\label{sec:intro}
Deep-learning methods require large training datasets to achieve high accuracy. Dataset distillation \cite{dataset_distillation} allows distilling large datasets into smaller ones so that training on the distilled dataset results in high accuracy. Concretely, dataset distillation methods synthesize the $K$ images-per-class (IPC) that are most relevant for the classification task. Dataset distillation has been very successful, achieving high accuracy with as little as a single image-per-class.

\begin{figure}[t]
\centering
\includegraphics[width=0.99\linewidth]{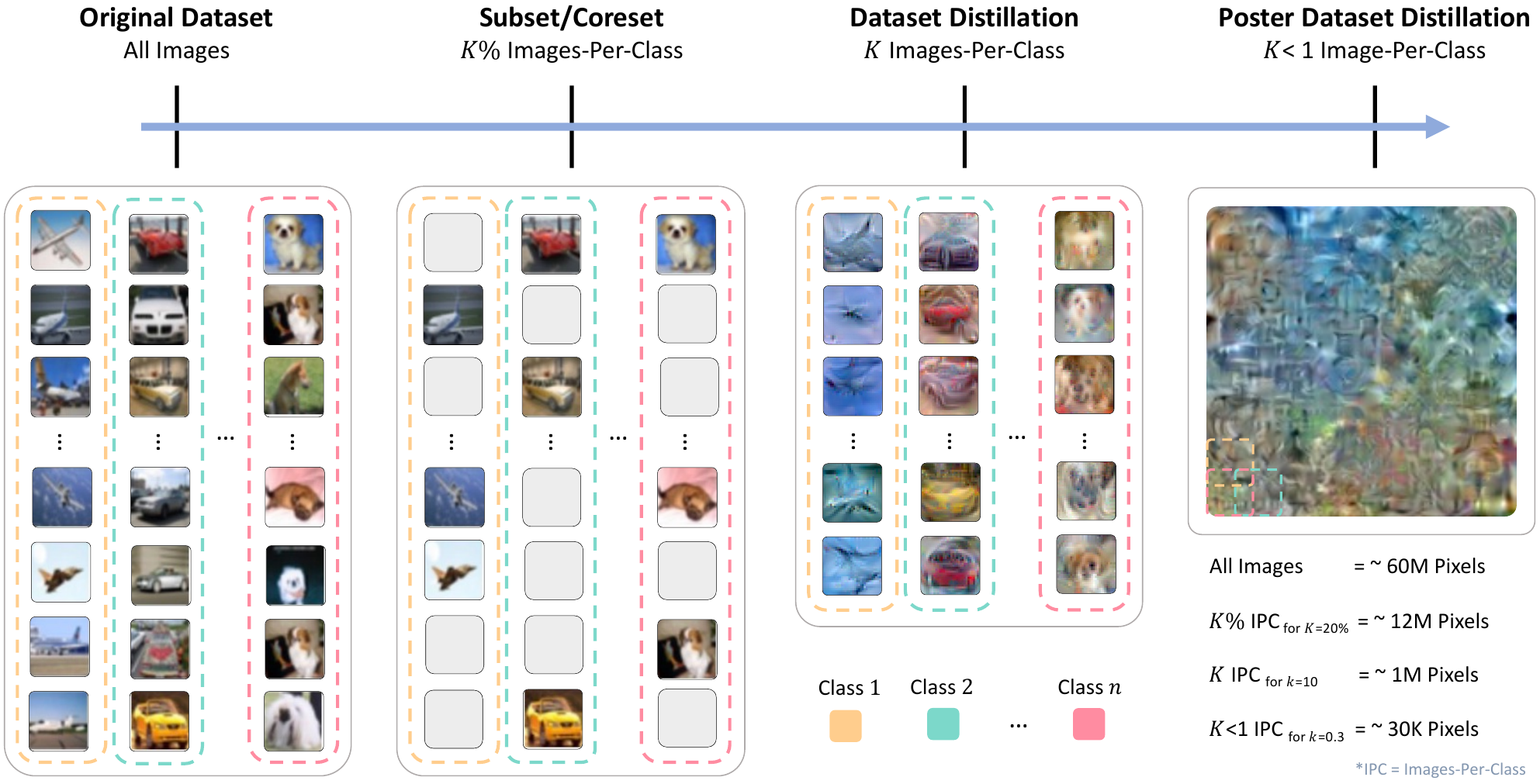}
\caption{\textit{\textbf{Dataset Compression Scale:}} We show increasingly more compressed methods from left to right. The original dataset contains all of the training data and does not perform any compression. Coreset methods select a subset of the original dataset, without modifying the images. Dataset distillation methods compress an entire dataset by synthesizing $K \in \mathbb{N}^{+}$ images-per-class (IPC). Our method, \textit{\textbf{Po}ster \textbf{D}ataset \textbf{D}istillation} (PoDD) distills an entire dataset into a single poster that achieves the same performance as $1$ IPC while using as little as $0.3$ IPC}

\label{fig:compression_scale}
\end{figure}

In this paper, we ask: \q{can we go lower than one image-per-class?} Existing dataset distillation methods are unable to do this as they synthesize one or more distinct images for each class. Assuming there are $n$ classes, such methods would require distilling at least $n$ images. On the other hand, using less than $1$ IPC implies that several classes share the same image, which current methods do not allow. We therefore propose \textit{\textbf{Po}ster \textbf{D}ataset \textbf{D}istillation} (PoDD), which distills an entire dataset into a single larger image, that we call a poster. The benefit of the poster representation is the ability to use patches that overlap between the classes.  We can set the size of the poster so it has significantly fewer pixels than in $n$ images, therefore enabling distillation with less than $1$ IPC. We find that a correctly distilled poster is sufficient for training a model with high accuracy. See \cref{fig:compression_scale} for an overview of different dataset compression methods.

To illustrate the idea of a poster, consider CIFAR-100\cite{cifar} where each image is of size $32\times 32$ pixels. Current methods synthesize images independently and thus must use at least $1$ IPC (see \cref{fig:1_vs_04_ipc}~(left)). Choosing $1$ IPC for CIFAR-100 entails using $100$ images, each of size $32 \times 32$ pixels. In contrast, PoDD synthesizes a single poster shared between all classes. During distillation, we optimize all the pixels so that a classifier trained on the resulting dataset achieves maximal accuracy. For example, to achieve $0.4$ IPC, we represent the entire dataset as a single poster of size $202 \times 202$ pixels (see \cref{fig:1_vs_04_ipc}~(right)). This has about the same number of pixels as $40$ images, each of size $32 \times 32$. The number of effective IPCs is therefore directly given by the size of the poster.

To distill a poster, we first initialize all pixels with random values. We transform the poster into a dataset in a differentiable way, by extracting overlapping patches, each with the same size as an image in the source dataset (e.g., $32 \times 32$ for CIFAR-100). During distillation, we optimize this set of overlapping poster patches and propagate the (overlapping) gradients from the distillation algorithm back to the poster. The optimization objective is to synthesize a poster such that a classifier trained on the dataset extracted from it will reach high classification accuracy. 
This process requires a label for every patch, we therefore propose  \textit{\textbf{Po}ster \textbf{D}ataset \textbf{D}istillation \textbf{L}abeling} (PoDDL), a method for poster labeling that supports both fixed and learned labels.

Since classes can now share pixels, their order within the poster matters as it implies which classes share pixels with each other. It is thus important to find the optimal ordering of classes on the poster. To this end, we propose \textit{\textbf{Po}ster \textbf{C}lass \textbf{O}rdering} (PoCO), an algorithm that uses CLIP \cite{clip} text embeddings to order the classes semantically and efficiently.

Overall, using less than $1$ IPC, PoDD can match or improve the performance that previous methods achieved with at least $1$ IPC. Indeed, sometimes PoDD can outperform competing methods using as low as $0.3$ IPC. Moreover, PoDD sets a new state-of-the-art for $1$ IPC on CIFAR-10, CIFAR-100, and CUB200. 

To summarize, our main contributions are:
\begin{enumerate}
    \item Proposing PoDD, a new dataset distillation setting for tiny, less than $1$ IPC budgets.
    \item Developing a method to perform PoDD that constitutes of a class ordering algorithm (PoCO) and a labeling strategy (PoDDL). 
    \item Performing extensive experiments that demonstrate the effectiveness of PoDD with as low as $0.3$ IPC and achieving a new $1$ IPC SoTA across multiple datasets. 
\end{enumerate}

\section{Related Works}
\label{sec:related_works}
Dataset distillation, introduced by \citet{dataset_distillation}, aims to compress an entire dataset into a smaller, synthetic one. The goal is that methods trained on the distilled dataset will achieve similar accuracy to a model trained on the original dataset. As highlighted by \cite{coresets_rebuffi2017,coresets_castro2018,coresets_intro}, dataset distillation shares similarities with coreset selection. Coreset selection identifies a representative subset of samples from the training set that can be used to train a model to the same accuracy. Unlike coreset selection, the generated synthetic samples of dataset distillation provide flexibility and improved performance through continuous gradient-based optimization techniques. Dataset distillation methods can be categorized into 4 main groups: i) \textit{Meta-Model Matching} \cite{dataset_distillation, kip, rfad, frepo, embarrassingly_simple} minimize the discrepancy between the transferability of models trained on a distilled data and those trained on the original dataset. ii) \textit{Gradient Matching }\cite{dc, dsa, dcc}, proposed by \citet{dc}, performs one-step distance matching between a network trained on the target dataset and the same network trained on the distilled dataset. This avoids unrolling the inner loop of Meta-Model Matching methods. iii) \textit{Trajectory Matching}\cite{mtt, tesla}, proposed by \citet{mtt}, focuses on matching the training trajectories of models trained on the target distilled dataset and the original dataset. iv) \textit{Distribution Matching}\cite{dm, cafe, kfs}, introduced by \citet{dm}, solves a proxy task via a single-level optimization, directly matching the distribution of the original dataset and the distilled dataset. See \cite{survey} for an in-depth explanation and comparisons of the various distillation methods. Common to all these methods is the use of at least one IPC. This paper proposes a method for distilling a dataset into less than $1$ IPC.

\section{Preliminaries}
\label{sec:preliminaries}
Many methods tackle dataset distillation as a bi-level optimization problem. In this setup, the inner loop essentially involves training a model with weights $\theta$ on a \textit{distilled} dataset $\mathcal{D}_{\text{syn}}$. The outer loop optimizes the pixels of the distilled dataset $\mathcal{D}_{\text{syn}}$, so a model trained on $\mathcal{D}_{\text{syn}}$ has the maximal accuracy on the \textit{original} dataset $\mathcal{D}$. Let $\mathcal{L}_{\mathcal{D}}(\theta)$ denote the average value of the objective function, of model $\theta$ on the dataset $\mathcal{D}$. Formally the optimization problem can be described as follows:
\begin{equation}
    \underbrace{
    \arg\min_{\mathcal{D}_{\text{syn}}}\hspace{0.01\linewidth}
    \mathcal{L}_{\mathcal{D}}\left( \theta^{*}\right)}_\text{Outer loop}
    \hspace{0.03\linewidth} \text{s.t} \hspace{0.03\linewidth}
    \theta^{*} = \underbrace{
    \arg\min_{\theta}\hspace{0.01\linewidth}
    \mathcal{L}_{\mathcal{D}_{\text{syn}}}(\theta)}_{\text{Inner loop}}
\end{equation}
We consider the case where the inner-loop optimization consists of $T_{end}$ SGD steps. The most common solution is backpropagation through time (BPTT) which unrolls the inner loop SGD optimization for $T_{end}$ steps. It then uses computationally demanding backpropagation calculation to compute the gradient of the loss with respect to the distilled dataset $\mathcal{D}_{\text{syn}}$,
\begin{equation}
    \underbrace{
    \arg\min_{\mathcal{D}_{\text{syn}}} \hspace{0.01\linewidth}
    \mathop{\mathbb{E}}_{\theta_{0} \sim P_{\theta}} \left[ \mathcal{L}_{\mathcal{D}}\left( \theta_{T_{end}}\right) \right]
    }_\text{Outer loop}
    \hspace{0.03\linewidth} \text{s.t} \hspace{0.03\linewidth}
    \underbrace{\theta_{t+1}\xleftarrow{}\theta_{t} - \eta \cdot \nabla_{\theta}\mathcal{L}_{\mathcal{D}_{\text{syn}}}(\theta_{t})}_{\text{Inner loop unrolling}}
\end{equation}

Backpropagating through $T_{end}$ timesteps is infeasible for large values of $T_{end}$, therefore many methods propose ways to reduce the computational costs. Here, we use RaT-BPTT \cite{embarrassingly_simple}, which computes the inner loop through a random number of SGD steps $T_{end} \sim Random[\Delta T, \Delta T+1, ..., T]$. It then approximates the gradient with respect to the distilled dataset by only backpropagating through the final $\Delta T<<T$ steps. We chose RaT-BPTT because it achieves the top performance across many dataset distillation benchmarks. 

\begin{figure}[t]
\vspace{-20pt}
    \centering
    \includegraphics[width=0.85\linewidth]{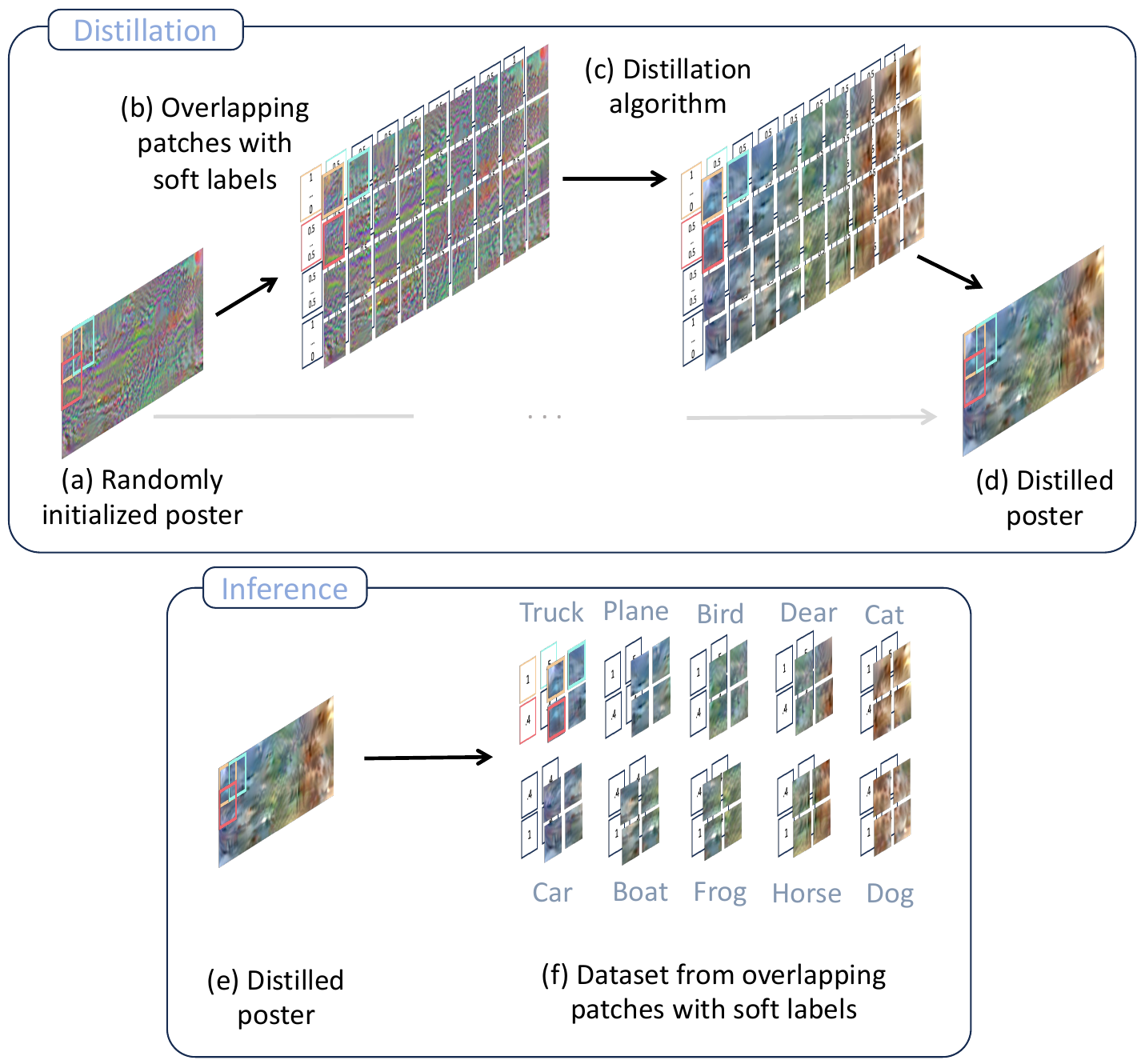}
    \caption{\textit{\textbf{PoDD Overview:}} We propose PoDD, a new dataset distillation setting for under 1 images-per-class. We start by initializing a random poster (a), during distillation, we optimize overlapping patches and soft labels (b-c). The final distilled poster has fewer pixels than the combined pixels of the individual images (d). During inference, we extract overlapping patches and soft labels from the distilled poster and use them to train a downstream model (e-f). PoDD achieves comparable or better accuracy to current methods while using as little as a third of the pixels}
    \label{fig:podd_overview}
\end{figure}

\section{PoDD: Poster Dataset Distillation}
\label{sec:podd_method}
Our goal is to perform dataset distillation with less than $1$ IPC. Our main insight is that sharing pixels between classes can be effective, as this would make better use of redundant pixels. Clearly, this requires more than one class to share an image (the pigeonhole principle \cite{pigeonhole_principle} states this formally).  In this section, we propose Poster Dataset Distillation (PoDD) which provides the methods to realize this idea.

\subsection{A Shared Poster Representation}
\label{sec:podd_def}

The key idea in this work is to distill an entire dataset into a single larger image that we call a poster. The poster can be of arbitrary height $d_h$ and width $d_w$, leading to a total of $d_h \times d_w$ pixels; posters with fewer pixels are said to be more \textit{compact}. Furthermore, we define a fixed procedure for converting a poster into a distilled dataset. Our procedure extracts multiple overlapping patches at a fixed stride; each extracted patch is equal in size to the original images (see \cref{fig:podd_overview} for a schematic of the procedure). We employ the same stride for both rows and columns and denote the total number of extracted patches by $p$. Consequently, this yields a dataset of images, some of which share pixels. 

We initialize the poster with standard Gaussian noise and optimize its pixels end-to-end through both the dataset expansion described above and the inner-loop bi-level optimization described in \cref{sec:preliminaries}. The size of the distilled dataset is measured by the total number of pixels in the poster, allowing us to compare with previous approaches that used one or more IPC. We provide pseudocode for PoDD in \cref{alg:podd}.

\begin{figure}[t]
    \centering
    \begin{minipage}{0.85\linewidth}
    \vspace{-25pt}
        \begin{algorithm}[H]
            \caption{\textit{\textbf{PoDD:}} Pseudocode using PoDDL learned labels}
            \label{alg:podd}
            \textbf{Input:} \textit{Alg}: Distillation algorithm.\\
            \textbf{Input:} $\mathcal{D}$: Dataset with classes $C$.\\
            \textbf{Input:} $d_h$, $d_w$: Poster size.\\
            \textbf{Input:} $p$: Number of overlapping patches.
            \begin{algorithmic}[1]
                \State $\mathcal{P} \sim \mathcal{N}(0,1)^{d_x \times d_w}$ \hspace{1.65cm} \color{teal}\#  Initialize a poster from Gaussian \color{black}
                \State  $O \leftarrow \text{PoCO}(C)$ \hspace{2.2cm} \color{teal}\# Initialize class order \color{black}
                \State $Y \leftarrow\text{PoDDL}(O, p)$ \hspace{1.55cm} \color{teal}\# Initialize distilled labels array \color{black}
                \For{each distillation step}
                    \State  $\mathcal{D}_{\text{syn}}: = \{ (p\text{, }l) \text{ overlapping patches and labels from } \mathcal{P} \text{ and }Y\}$ 
                    \State $\mathcal{P} \text{, } Y \leftarrow \textit{Alg}(\mathcal{D}, \mathcal{D}_{\text{syn}})$ \hspace{0.8cm} \color{teal}\# Distill one step \color{black}
                \EndFor
                \State Return $\mathcal{P}$, $Y$
            \end{algorithmic}
        \end{algorithm}
    \end{minipage}
\end{figure}

\subsection{PoCO: Poster Class Ordering}
\label{sec:poco_class_order}

\begin{figure}[b]
    \begin{minipage}{.54\textwidth}
        \begin{algorithm}[H]
            \caption{\textbf{\textit{PoCO:}} Pseudocode for PoCO class ordering}
            \label{alg:poco}
            \textbf{Input:} $\mathcal{E}$: CLIP Text encoder. \\
            \textbf{Input:} $o_x, o_y$: Class grid dimensions. \\
            \textbf{Input:} $C$: Class list of the target dataset.
            \begin{algorithmic}[1]
                \State $O \leftarrow 0^{o_x \times o_y}$  
                \State $O_{[0, 0]} \leftarrow C_{[0]}$ \hspace{0cm} 
                \State $C \leftarrow C \backslash \{C_{[0]}\}$ 
                \State $G \leftarrow$ Distance matrix using $\mathcal{E}$ embeddings
                \For{$i, j$ in $o_x, o_y$ (Zigzag traverse)}
                    \State $ngbr \leftarrow $ \{$(i, j)$ neighboring classes in $O\}$
                    \State $c \leftarrow C[\arg\min_{ngbr}(G)]$
                    \State $O_{[i, j]} \leftarrow c$
                    \State $C \leftarrow C \backslash \{c\}$
                \EndFor
                \State \textbf{return} $O$
            \end{algorithmic}
        \end{algorithm}
    \end{minipage}
    \hspace{0.01\textwidth}
    \begin{minipage}{.44\textwidth}
        \centering
        \includegraphics[width=0.50\linewidth]{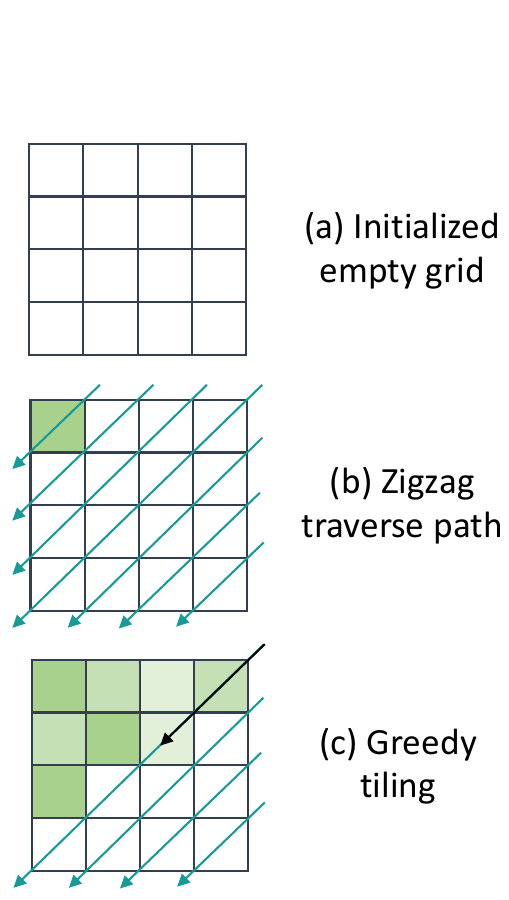}
        \caption{\textit{\textbf{PoCO Tiling:}} (a) Initialize empty grid $O$, (b) Traverse $O$ in a Zigzag order, (c) Tile $O$ in a greedy manner using CLIP text embeddings}
        \label{fig:poco_zigzag}
    \end{minipage}%
\end{figure}

The poster representation relies on shared pixels between neighboring classes. To maximize the effectiveness of the shared pixels, it is therefore important to establish the optimal structure of neighboring classes. We hypothesize that a poster would be more effective when pixels are shared between semantically related classes (e.g., Man, Woman, Boy vs. Plate, Tree, Bus). 

We propose a method for approximating the optimal class neighborhood structure. First, we extract an embedding for each class name using the CLIP \cite{clip} text-encoder  $\mathcal{E}$. Given the set of embeddings, we calculate the pairwise distance matrix between all classes. Using this pairwise class distance, we design a greedy procedure for placing the classes on a rectangular grid $O^{o_x\times o_y}$, denoting the spatial positions of classes. Let $C$ be the set of classes and $C_p$ be the set of already placed classes. 
We traverse the grid in a Zigzag order \cite{zigzag}, and at each step place a class as follows: 
\begin{equation*}
    O_{i,j}=\min_{c \in C\backslash C_p }\left( \sum_{m \in \{ O_{i-1,j}, O_{i,j-1}\}} \left( 1 - \frac{\mathcal{E} (m) \cdot \mathcal{E} (c)}{||\mathcal{E} (m)|| \cdot ||\mathcal{E} (c)||}\right)\right), \quad i\in[o_x], \hspace{0.2cm} j \in [o_y]
\end{equation*}
Intuitively, at each step, we place a remaining class that has the lowest distance from all its already-placed neighbors. We visualize this Zigzag traversal in \cref{fig:poco_zigzag} and summarize the PoCO algorithm in \cref{alg:poco}, in \cref{fig:PoCO_clusters} we show an example PoCO tiling for CIFAR-100.

\subsection{PoDDL: Poster Dataset Distillation Labeling}
\label{sec:poddl_labeling}
\begin{figure}[t]
\centering
\includegraphics[width=.9\linewidth]{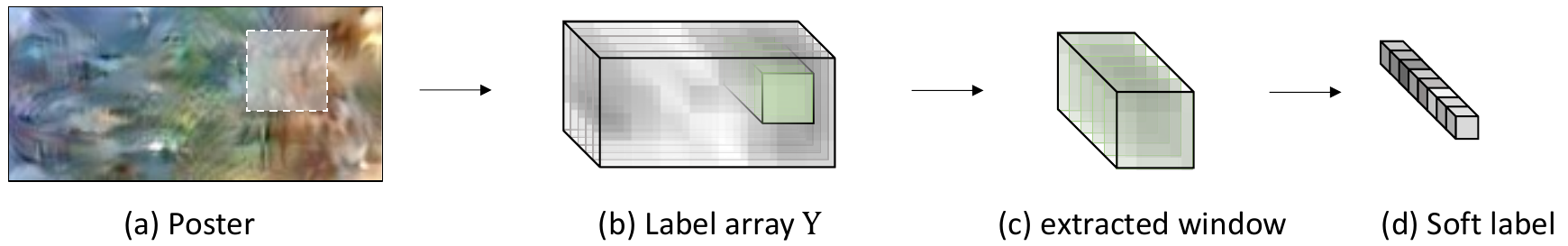}
\caption{\textit{\textbf{PoDDL Extraction:}} Each poster patch has a corresponding patch in the label array (a-b). We compute the poster patch label by extracting a patch along the channels of the label array (c). To obtain the final soft label for a given poster patch, we pool and normalize the extracted label window, resulting in a soft label vector (d). PoDDL supports both fixed and learned labels}
\label{fig:poddl_overview}
\end{figure}

Having initialized the poster and the class order matrix $O$, we now describe our labeling strategy. Previous approaches use one or more images-per-class, hence, they can simply assign a single label per image. However, using a single label for the entire poster is not a good option, instead, we assign a soft label \cite{soft_labels} vector to each overlapping patch. We therefore design a poster-oriented soft labeling strategy that supports both fixed and learned labels, see \cref{fig:poddl_overview} for an overview.

\noindent \textbf{Fixed labels.} We upsample the class order matrix $O$ to the size of the poster $d_h \times d_w$. For each overlapping patch, we extract its corresponding class label window. We compute its majority class and use it as the one-hot label for the patch. In the case of ties, we use a soft label with equal probabilities for the majority classes. 

\noindent \textbf{Learned labels.} As our method extracts an arbitrary number of overlapping patches, learning a soft label for each patch would require more parameters than previous approaches. To keep the the number of parameters constant, we learn a parameter tensor of the same size as previous works, and interpolate it to each overlapping window.

Concretely, we learn a label tensor $Y$ of size $o_x \times o_y \times n$ and spatially upscale it to the shape of the poster using nearest neighbor interpolation. The final size of $Y$ is $d_h \times d_w \times n$. For each overlapping patch, we extract the corresponding label window and average pool it. To achieve a valid label distribution, we $L_1$ normalize the resulting vector. We use this vector as the learned soft label of the window. After each gradient step, we clip negative values of $Y$ to zero to avoid negative probabilities. Unlike the fixed labels, the learned labels are optimized alongside the distillation process.

\section{Experiments}
\label{sec:experiments}
\subsection{Experimental Setting}
\label{sec:experimental_setting}
\noindent \textbf{Datasets.} We evaluate PoDD on four datasets commonly used to benchmark dataset distillation methods: i) \textit{CIFAR-10:} $10$ classes, $50k$ images of size $32 \times 32 \times 3$ \cite{cifar}. ii) \textit{CIFAR-100:} $100$ classes, $50k$ images of size $32 \times 32 \times 3$ \cite{cifar}. iii) \textit{CUB200:} $200$ classes, $6k$ images of size $32 \times 32 \times 3$ \cite{cub200}. iv) \textit{Tiny-ImageNet:} $200$ classes, $100k$ images of size $64 \times 64 \times 3$ \cite{tiny_imagenet}. 

\noindent \textbf{Distillation Method.} For the dataset distillation algorithm, we use RaT-BPTT \cite{embarrassingly_simple}, a recent method that achieved SoTA performance on CIFAR-10, CIFAR-100, and CUB200 by a wide margin. In particular, RaT-BPTT's architecture uses three convolutional layers for $32 \times 32$ datasets and four layers for $64 \times 64$ datasets.
\noindent \textbf{Baselines.} Our baselines can be divided into two groups: i) \textit{Inner-Loop:} BPTT \cite{bptt}, empirical feature kernel (FRePO) \cite{frepo}, and reparameterized convex implicit gradient (RCIG) \cite{rcig}, and ii) \textit{Modified Objectives:} gradient matching with augmentations (DSA) \cite{dsa}, distribution matching (DM) \cite{dm}, trajectory matching (MTT) \cite{mtt}, flat trajectory distillation (FDT) \cite{ftd}, and TESLA \cite{tesla}. We use the results reported by the baselines. 

\noindent \textbf{Evaluation.} Following the protocol of \cite{dsa,bptt}, we evaluate the distilled poster using a set of $8$ different randomly initialized models with the same ConvNet \cite{convnet} architecture used by DSA, DM, MTT, RaT-BPTT, and others. The architecture includes convolutional layers of $128$ filters with kernel size $3 \times 3$ followed by instance normalization\cite{instance_normalization}, ReLU \cite{relu} activation, and an average pooling layer with kernel size $2 \times 2$ and stride $2$. We report the mean and standard deviation across the $8$ random models. We evaluate two IPC regimes: 

\textit{Less than one IPC.} We compute the results for IPCs within the range: $K \in [0.3,0.4,...,1)$. Our criterion for success is if PoDD with $K$ IPC performs comparably or better than RaT-BPTT with $1$ IPC. To comply with previous baselines, we used PoDDL, which optimized the same number of label parameters as previous methods with $1$ IPC. 

\textit{One IPC.} To properly compare PoDD with existing methods we also evaluate it with the same total number of pixels used by our baselines (1 IPC). Since PoDD still uses overlapping patches, this evaluates the impact of compressing inter-class redundancies in the shared pixels, this is compared to the baselines which are unable to do so.

\noindent \textbf{Implementation Details} 
We use the same distillation hyper-parameters used by RaT-BPTT \cite{embarrassingly_simple} except for the batch sizes. To fit the distillation into a single GPU (we use an NVIDIA A40), we use the maximal batch size we can fit into memory for a given dataset (see exact breakdown below). Since the optimization is bi-level, distillation methods have two batch types, one for the distilled data which we denote by $bs_d$, and one for the original dataset which we denote by $bs$.

In addition to the $K$ IPC parameter, we need to control the degree of patch overlap. In other words, given a dataset with $n$ classes and after fixing $K$ (i.e., once we fix the size of the poster), we need to decide on the number of overlapping patches to divide the poster into. As the number of classes and image resolution varies between the datasets, we use a different $p$ for each one. Concretely, we use: i) \textit{CIFAR-10}: $p=96_{(16 \times 6)}\text{ patches},\hspace{0.05cm} bs_d=96,\hspace{0.05cm} bs=5000,\hspace{0.05cm} 4k$ epochs. ii) \textit{CIFAR-100}: $p=400_{(20 \times 20)}\text{ patches},\hspace{0.05cm} bs_d=50,\hspace{0.05cm} bs=2000,\hspace{0.05cm} 2k$ epochs.  iii) \textit{CUB200}: $p=1800_{(60 \times 30)}\text{ patches},\hspace{0.05cm} bs_d=200,\hspace{0.05cm} bs=3000,\hspace{0.05cm} 8k$ epochs. iv) \textit{Tiny-ImageNet}: $p=800_{(40 \times 20)}\text{ patches},\hspace{0.05cm} bs_d=30,\hspace{0.05cm} bs=500,\hspace{0.05cm} 500$ epochs.     

We use the learned labels variant of PoDDL for all of our experiments except for CIFAR-10 with $K\in[0.7, 0,8, 0.9, 1.0]$ IPC in which we use the fixed labels variant (the learned labels did not provide additional benefit in these cases). We use a learning rate of $0.001$ for CIFAR-10, CIFAR-100, and CUB200. To fit Tiny-ImageNet into a single GPU we use a much smaller batch size and a learning rate of $0.0005$.

\begin{table*}[t]
\caption{\textit{\textbf{Less than One Image-Per-Class:}} 
PoDD with less than one IPC (image-per-class) often outperforms state-of-the-art methods with 1-IPC; in some cases with as little as $0.3$ IPC. In \color{drop}{(red)}\color{black}, the relative performance drop compared to the 1 IPC results. In \textbf{bold}, the lowest IPC for which PoDD beats the current SoTA
}
\centering
\begin{tabular}{l|c|c|c|c|c}
Method & IPC $\downarrow$ & CIFAR-10  $\uparrow$ & CIFAR-100 $\uparrow$ & CUB200 $\uparrow$ & T-ImageNet $\uparrow$\\
\midrule

RaT-BPTT \cite{embarrassingly_simple} & 1.0 & 53.2\std{0.7} & 35.3\std{0.4} & 13.8\std{0.3} & 20.1\std{0.3} \\
PoDD (Ours) & 1.0 & 59.1\std{0.5} & 38.3\std{0.2} & 16.2\std{0.3} & 20.0\std{0.3} \\
\midrule
\multirow{7}{*}{PoDD (Ours)}& 0.9 &
                            $58.4\std{0.5}_{\color{drop}(1\%)}$ & $37.4\std{0.2}_{\color{drop}(2\%)}$ &
                            $15.2\std{0.4}_{\color{drop}(6\%)}$ & $19.5\std{0.2}_{\color{drop}(2\%)}$ \\
                            & 0.8 & 
                            $56.7\std{0.7}_{\color{drop}(4\%)}$ & $37.3\std{0.1}_{\color{drop}(3\%)}$ & $15.6\std{0.3}_{\color{drop}(4\%)}$ & $19.0\std{0.2}_{\color{drop}(5\%)}$ \\
                            & 0.7 & 
                            $\textbf{54.6}\std{0.5}_{\color{drop}(8\%)}$ & $37.0\std{0.2}_{\color{drop}(3\%)}$ & $15.0\std{0.3}_{\color{drop}(8\%)}$ & $18.6\std{0.1}_{\color{drop}(7\%)}$ \\
                            & 0.6 & 
                            $50.6\std{0.3}_{\color{drop}(15\%)}$ & $36.6\std{0.3}_{\color{drop}(5\%)}$ & $15.1\std{0.2}_{\color{drop}(7\%)}$ & $18.8\std{0.2}_{\color{drop}(6\%)}$ \\
                            & 0.5 & 
                            $49.5\std{0.5}_{\color{drop}(16\%)}$ & $36.0\std{0.3}_{\color{drop}(6\%)}$ & $15.0\std{0.3}_{\color{drop}(8\%)}$ & $18.7\std{0.1}_{\color{drop}(7\%)}$ \\
                            & 0.4 & 
                            $47.1\std{0.4}_{\color{drop}(20\%)}$ & $\textbf{35.7}\std{0.2}_{\color{drop}(7\%)}$ & $15.0\std{0.2}_{\color{drop}(7\%)}$ & $18.4\std{0.2}_{\color{drop}(8\%)}$ \\
                            & 0.3 & 
                            $42.3\std{0.3}_{\color{drop}(28\%)}$ & $34.7\std{0.2}_{\color{drop}(10\%)}$ & $\textbf{14.8}\std{0.5}_{\color{drop}(9\%)}$ & $18.4\std{0.1}_{\color{drop}(8\%)}$ \\
                            
\midrule
\begin{tabular}[l]{@{}l@{}}Full Dataset \\(No Dist.)\end{tabular}& \begin{tabular}[c]{@{}c@{}}All\\Images\end{tabular}   & 83.5\std{0.2} & 55.3\std{0.3} & 20.1\std{0.3} & 37.6\std{0.5} \\
\end{tabular}
\label{tab:ipc_lt_1}
\end{table*}

\subsection{Results}
\label{sec:results}
\noindent \textbf{Less than one IPC.} We now test our initial question: \q{can we go lower than one image-per-class?} Using PoDD, we show that across all four datasets, we can go much lower than $1$ IPC and for $3$ out of the $4$ datasets even maintain on-par performance to the SoTA baseline. As hypothesized, using a poster that shares pixels between multiple classes allows us to reduce redundancies between classes in the distilled patches (See \cref{tab:ipc_lt_1}). This effect is intensified when distilling datasets with a large number of classes, e.g., for CIFAR-100 we can use $0.4$ IPC and for CUB200 we can use as little as $0.3$ IPC and still outperform the baseline method.

\noindent \textbf{One IPC.} Having shown the feasibility of distilling a dataset into less than one IPC, we now quantitatively evaluate the benefit of the poster representation. To this end, we use the $1$ IPC setting which allows us to decouple the pixel count from the pixel sharing. Essentially, we are investigating whether the pixel-sharing in our poster can boost performance, even when the number of pixels matches our baseline. Our method outperforms the state-of-the-art for CIFAR-10, CIFAR-100, and CUB200, setting a new SoTA for $1$ IPC dataset distillation (See \cref{tab:ipc_1}). 

\begin{table*}[t]
\caption{\textit{\textbf{One Image-Per-Class:}} Performance of PoDD under the $1$ image-per-class (IPC) setting compared to SoTA dataset 
distillation methods across $4$ datasets. PoDD sets a new SoTA for CIFAR-10, CIFAR-100, and CUB200. On Tiny-ImageNet, PoDD achieves comparable results to the underlying distillation method it uses (RaT-BPTT)
}
\centering
\begin{tabular}{cl|c|c|c|c|c}
& Method & CIFAR-10 $\uparrow$ & CIFAR-100 $\uparrow$ & CUB200 $\uparrow$ & T-ImageNet $\uparrow$ & Average $\uparrow$ \\
 \cmidrule{2-7}
 \multirow{4}{*}{\rotatebox[origin=c]{90}{\color{gray}Inner Loop}} & BPTT \cite{bptt} & 49.1\std{0.6} & 21.3\std{0.6} & - & - & -\\
 & FRePO \cite{frepo} & 45.6\std{0.1} & 26.3\std{0.1} & - & 16.9\std{0.1} & - \\
 & RCIG \cite{rcig} & 49.6\std{1.2} & 35.5\std{0.7} & - & \textbf{22.4}\std{0.3} & - \\
& RaT-BPTT \cite{embarrassingly_simple} & 53.2\std{0.7} & 35.3\std{0.4} & 13.8\std{0.3} & 20.1\std{0.3} & 30.6\std{0.4} \\
\cmidrule{2-7}
 \multirow{5}{*}{\rotatebox[origin=c]{90}{\begin{tabular}[c]{@{}c@{}}\color{gray}Modified \\\color{gray}Objectives\end{tabular}}}  & DSA \cite{dsa} & 28.8\std{0.7} & 13.9\std{0.3} & 1.3\std{0.1} & 6.6\std{0.2} & 12.7\std{0.3} \\
 & DM \cite{dm} & 26.0\std{0.8} & 11.4\std{0.3} & 1.6\std{0.1} & 3.9\std{0.2} & 10.8\std{0.3} \\
 & MTT \cite{mtt} & 46.3\std{0.8} & 24.3\std{0.3} & 2.2\std{0.1} & 8.8\std{0.3} & 20.4\std{0.4}\\
& FTD \cite{ftd} & 46.8\std{0.3} & 25.2\std{0.2} & - & 10.4\std{0.3} & - \\
& TESLA \cite{tesla} & 48.5\std{0.8} &  24.8\std{0.4} & - & - & - \\
\cmidrule{2-7}
& PoDD (Ours) & \textbf{59.1}\std{0.5} & \textbf{38.3}\std{0.2} & \textbf{16.2}\std{0.3} & 20.0\std{0.3} & \textbf{33.4}\std{0.3}\\
\cmidrule{2-7}
& \begin{tabular}[l]{@{}l@{}}Full Dataset \\(No Distillation)\end{tabular} & 83.5\std{0.2} & 55.3\std{0.3} & 20.1\std{0.3} & 37.6\std{0.5} &  49.1\std{0.3}\\
\end{tabular}
\label{tab:ipc_1}
\end{table*}

\subsection{Ablations}

\noindent \textbf{Class order ablation.} 
\label{sec:class_order}
We ablate the impact of the class ordering on the performance of PoDD on CIFAR-10. We first compute the distillation performance after $250$ distillation steps with $0.3$ IPC for $5$ random class orderings. The score of each ordering is the inverse of the sum of distances between all neighboring class pairs. The distance matrix is defined in the same way as in PoCO i.e., using the embeddings of the CLIP text encoder. We find that the class ordering can indeed impact the performance of the distilled poster, with a correlation coefficient of $0.76$ between the score of the ordering and the accuracy the distilled poster achieves. This correlation motivates PoCO's search for the optimal class ordering.

\begin{figure}[t]
\centering
\begin{minipage}{.5\textwidth}
  \centering
  \begin{table}[H]
    \centering
    \begin{tabular}{c@{\hskip10pt}c}
      \begin{tabular}[c]{@{}c@{}}Number of \\Patches ${(o_{x} \times o_{y})}$\end{tabular}  & \begin{tabular}[c]{@{}c@{}}Test Accuracy \\$500_{epochs}$ \end{tabular}  \\
      \midrule
      $10_{(5 \times 2)}$ & $45.15 \%$\\
      $24_{(8 \times 3)}$ & $47.28 \%$\\
      $40_{(10 \times 4)}$ & $56.77 \%$\\
      $60_{(12 \times 5)}$ & $54.14 \%$\\
      $96_{(16 \times 6)}$ & $56.73 \%$\\
      $126_{(18 \times 7)}$ & $55.55 \%$\\
      $160_{(20 \times 8)}$ & $57.61 \%$\\
    \end{tabular}
  \end{table}
\end{minipage}%
\begin{minipage}{.5\textwidth}
  \centering
  \includegraphics[width=0.8\linewidth]{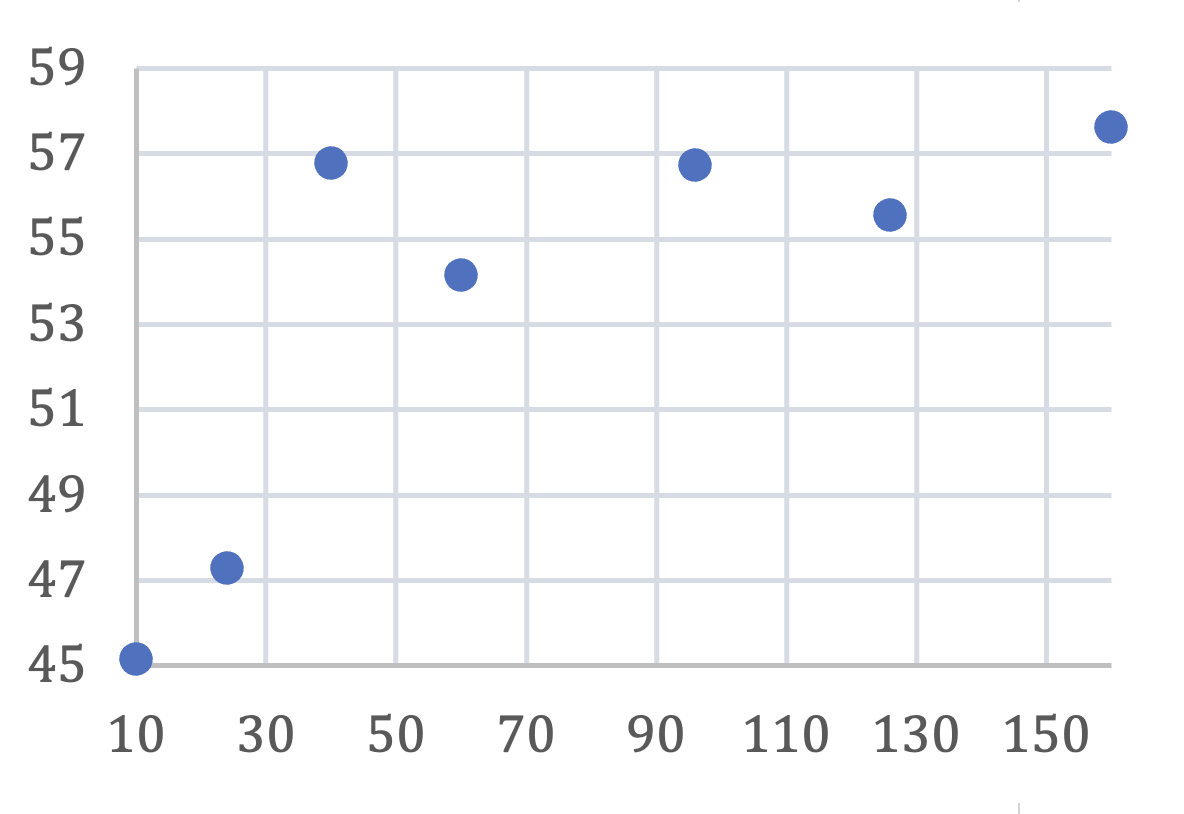}
\end{minipage}
\caption{\textit{\textbf{Number of Patches Ablation:}} We ablate the effect of the patch overlap on the accuracy (CIFAR-10, $1$ IPC, $500$ distillation steps). Using $10$ patches, which is the number of classes (i.e., no patch overlap) results in the lowest accuracy. When increasing the number of patches beyond $24$, the results improve significantly}
\label{fig:ablation_patch_num}
\end{figure}

\noindent \textbf{Patch number ablation.}
We ablate the role of the amount of overlap between patches of the poster (i.e., the number of patches for a given poster size and dataset). To study this, we use CIFAR-10 at $1$ IPC, we run PoDD for $500$ steps multiple times, each with a progressively increasing number of patches. As can be seen in \cref{fig:ablation_patch_num}, using the same number of patches as the number of classes (i.e., no overlap between patches) results in the lowest score; this is expected as this is exactly the RaT-BPTT baseline. When increasing the number of patches, we observe that beyond a certain patch number threshold, the results improve drastically. This demonstrates the significance of the poster representation and the use of overlapping patches. Since the number of patches has a direct effect on the distillation time and the training time of downstream models, we use a small number of patches for the larger datasets and a larger number of patches for the smaller datasets.

\section{Discussion and Future Work}
\label{sec:discussion}
Beyond the exciting result of distilling a dataset into less than one image, PoDD presents a new setting and distillation representation that opens up new and intriguing research problems.

\begin{figure}[t]
\centering
\includegraphics[width=\linewidth]{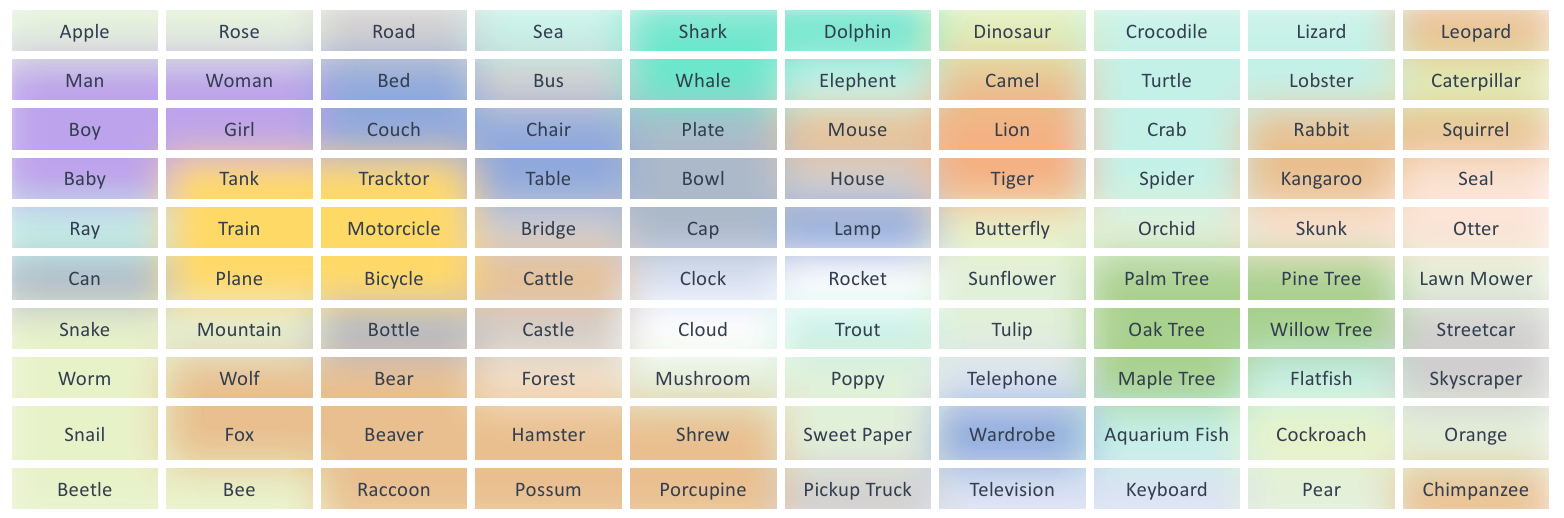}
\caption{\textit{\textbf{PoCO Ordering:}} An example output of PoCO for CIFAR-100. The classes are separated into semantically meaningful classes (e.g., trees, humans, vehicles). We colored semantically related clusters manually}
\label{fig:PoCO_clusters}
\end{figure}

\noindent \textbf{Class ordering algorithm.}
\label{sec:discussion_class_order}
As shown in \cref{sec:class_order}, the ordering of the classes within a poster is strongly correlated with the performance of the distilled poster. We proposed PoCO, a greedy algorithm for choosing a class ordering. In \cref{fig:PoCO_clusters} we show an example ordering for CIFAR-100, as can be seen, the classes are separated into semantically meaningful classes (e.g., trees, humans, vehicles). However, PoCO does not always yield a perfect ordering, e.g., the leopard may not fit well in the top right corner. It might fit better next to the lion and the tiger ($4$ left and $2$ down). Indeed, other ordering strategies may be better suited for the distillation task, e.g., a photometric-based ordering that uses the color of the images or an ordering that uses the image semantics. Further investigation of alternative ordering methods is left for future work.

\begin{figure}[t]
\centering
    \includegraphics[width=0.9\linewidth]{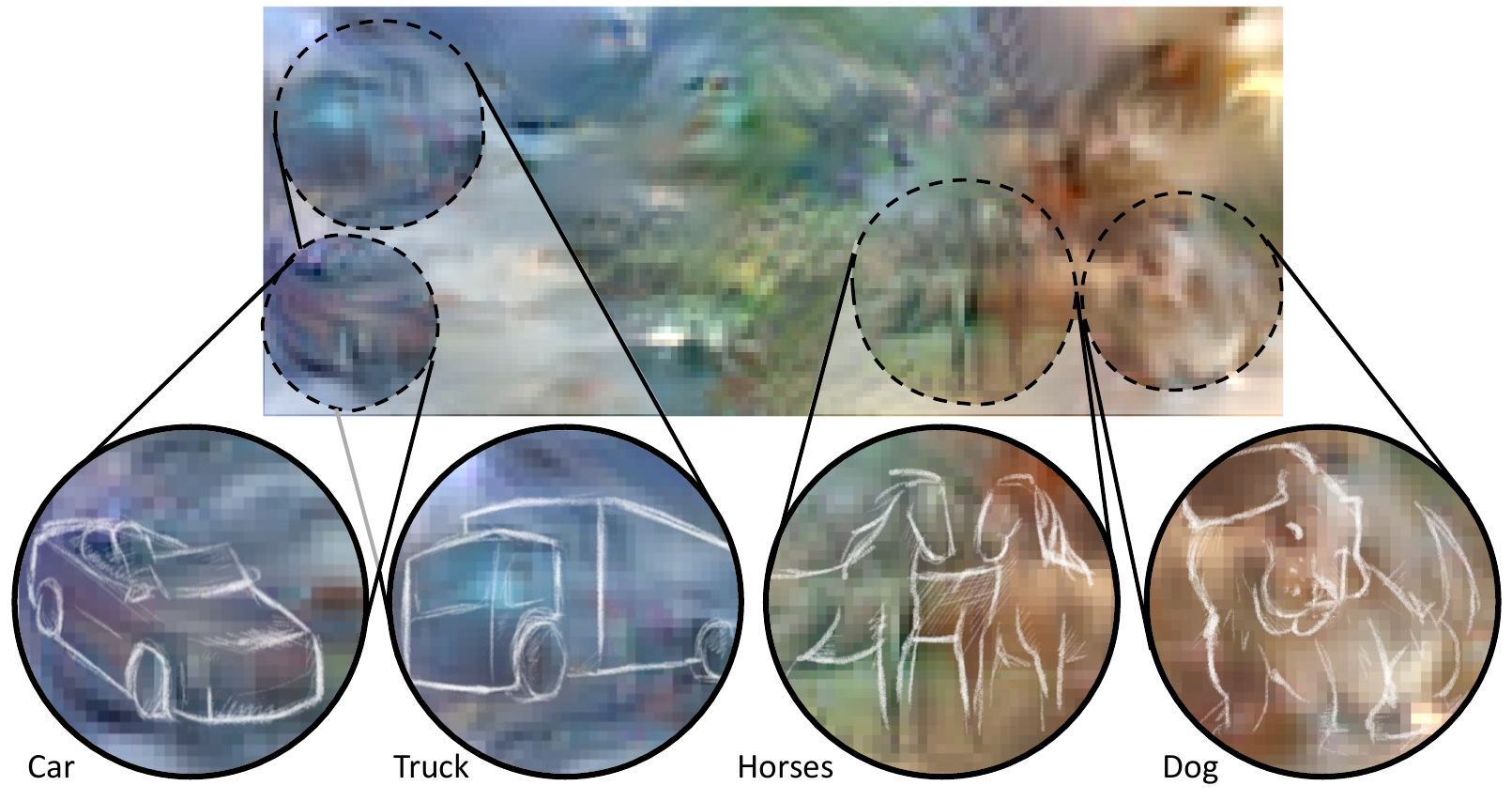}
    \caption{\textit{\textbf{Distilled Poster Semantics:}} We illustrate some of the semantics captured by a CIFAR-10, $1$ IPC poster by sketching over the distilled poster. The poster is of dimension $5 \times 2$, with the top row containing: Truck, Plane, Bird, Deer, Cat, and the bottom row containing Car, Boat, Frog, Horse, Dog. We can see that the pixels shared between classes exhibit a smooth transition between colors}
    \label{fig:global_local_cifar10}
\end{figure}

\noindent \textbf{Other IPC values.} In this work, we focused on $\leq 1$ IPC, however, it is possible to extend PoDD to values exceeding $1$ IPC. In a preliminary investigation, we found that $1\leq$ IPC achieves SoTA results for some of the datasets. However, we expect that leveraging the full potential of $1\leq$ IPC will require new labeling and class ordering algorithms.

\begin{figure}[t]
\centering
    \includegraphics[width=\linewidth]{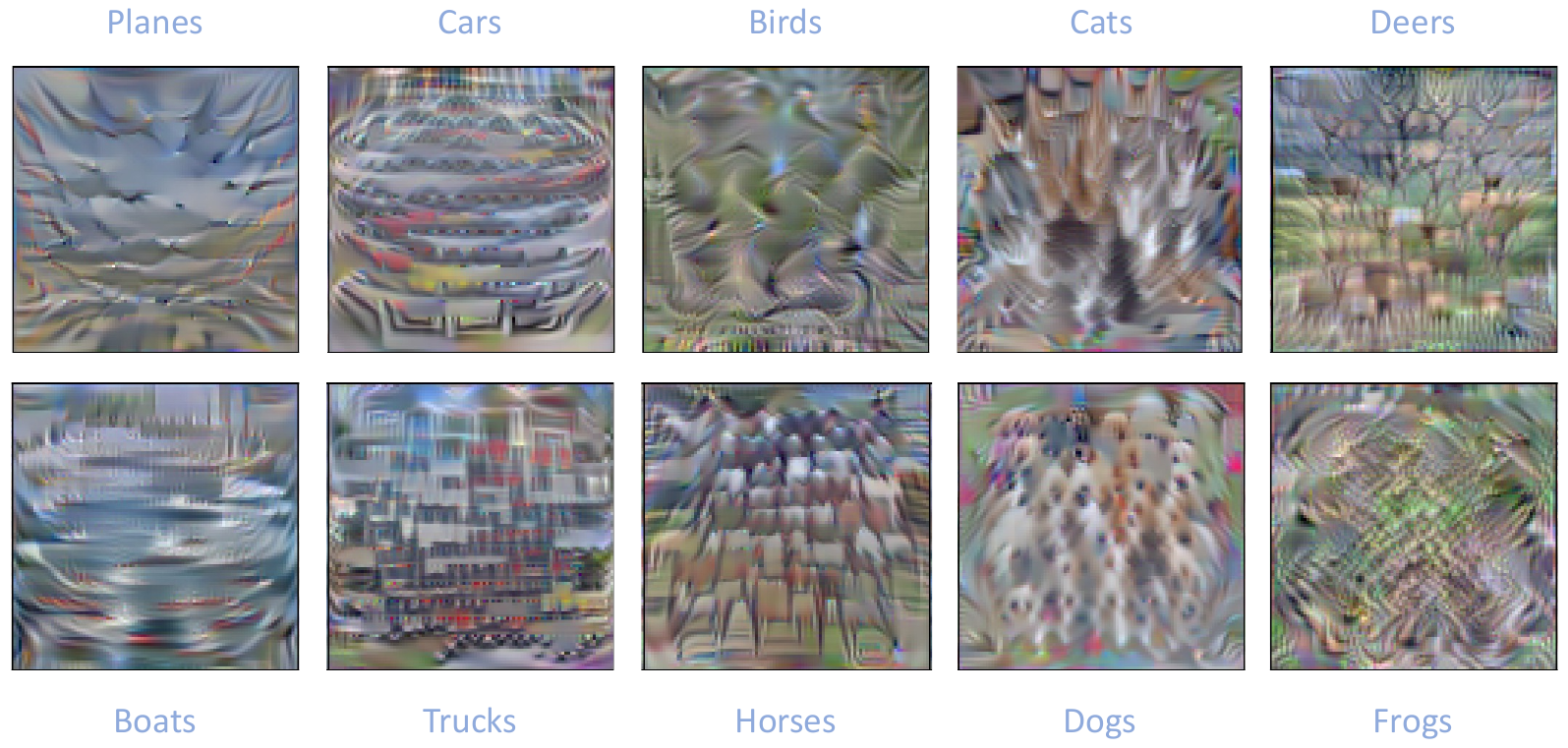}
    \caption{\textit{\textbf{Global and Local Semantics:}} We train a CIFAR-10 variant of PoDD with $10$ IPC and a separate \textit{per class} poster. The local semantics are well preserved, showing multiple modalities per class, e.g., different colors of cars, poses of animals, and locations of the planes. Moreover, some of the classes demonstrate global scale semantics, e.g., the planes have sky on the top and grass on the bottom}
    \label{fig:tesla_podd}
\end{figure}

\noindent \textbf{Global and local scale semantic results.} We investigate whether PoDD can produce distilled posters that exhibit both local and global semantics. We found that in the case of $1$ IPC, both local and global semantics are present, but are hard to detect. For example, in \cref{fig:global_local_cifar10} we illustrate some of the captured semantics by sketching over the poster. To further explore this idea, we tested \cite{tesla} with a CIFAR-10 variant of PoDD where we use $10$ IPC and distilled a poster \textit{per class}. Each poster now represents a single class and overlapping patches are always from the same class. To enable this, we retrofit PoDDL by increasing the size of $Y$ proportionally by $\sqrt{K}$, allowing it to operate with $1 < K$. As seen in \cref{fig:tesla_podd}, the method preserves the local semantics and shows multiple modalities from each class. Moreover, some of the posters also demonstrate global semantics, e.g., the planes have the sky on the top and the grass on the bottom. 

\noindent \textbf{Patch Augmentations}
Throughout this work we use the extracted patches with no modifications. However, performing spatial augmentations (e.g., scale, rotation) on the distilled patches during the distillation process may be beneficial. Another option is to create a cyclic poster where patches near the border of the poster are wrapped around.

\section{Conclusion}
\label{sec:conclusion}
In this work, we propose poster dataset distillation (PoDD), a new dataset distillation setting for tiny, less than $1$ image-per-class budgets. We develop a method to perform PoDD and present a strategy for ordering the classes within the poster. We demonstrate the effectiveness of PoDD by achieving on-par SoTA performance with as low as $0.3$ IPC and by setting a new $1$ IPC SoTA performance for CIFAR-10, CIFAR-100, and CUB200.

\section{Broader Impact}
Our work demonstrates the potential for PoDD to reduce the environmental footprint of deep learning by achieving higher compression rates for image-based datasets. This innovation not only lowers storage requirements but also leads to reduced training times, fostering more sustainable deep-learning research practices. By introducing the pixels-per-dataset approach to dataset distillation, we encourage the development of more efficient and resource-conscious practices.

However, it's essential to acknowledge that, like any machine learning project, our work may have various societal consequences. While we believe these should be considered, we do not feel it necessary to delve into specific societal impacts in this statement.

\clearpage
\bibliographystyle{ieeenat_fullname}
\bibliography{egbib}

\begin{thebibliography}{30}
\providecommand{\natexlab}[1]{#1}
\providecommand{\url}[1]{\texttt{#1}}
\expandafter\ifx\csname urlstyle\endcsname\relax
  \providecommand{\doi}[1]{doi: #1}\else
  \providecommand{\doi}{doi: \begingroup \urlstyle{rm}\Url}\fi

\bibitem[Castro et~al.(2018)Castro, Mar{\'\i}n-Jim{\'e}nez, Guil, Schmid, and Alahari]{coresets_castro2018}
Francisco~M Castro, Manuel~J Mar{\'\i}n-Jim{\'e}nez, Nicol{\'a}s Guil, Cordelia Schmid, and Karteek Alahari.
\newblock End-to-end incremental learning.
\newblock In \emph{Proceedings of the European conference on computer vision (ECCV)}, pages 233--248, 2018.

\bibitem[Cazenavette et~al.(2022)Cazenavette, Wang, Torralba, Efros, and Zhu]{mtt}
George Cazenavette, Tongzhou Wang, Antonio Torralba, Alexei~A. Efros, and Jun-Yan Zhu.
\newblock Dataset distillation by matching training trajectories.
\newblock In \emph{Proceedings of the IEEE/CVF Conference on Computer Vision and Pattern Recognition (CVPR)}, pages 4750--4759, 2022.

\bibitem[Chai et~al.(2017)Chai, Gan, Chen, and Zhang]{zigzag}
Xiuli Chai, Zhihua Gan, Yiran Chen, and Yushu Zhang.
\newblock A visually secure image encryption scheme based on compressive sensing.
\newblock \emph{Signal Processing}, 134:\penalty0 35--51, 2017.

\bibitem[Cui et~al.(2023)Cui, Wang, Si, and Hsieh]{tesla}
Justin Cui, Ruochen Wang, Si Si, and Cho-Jui Hsieh.
\newblock Scaling up dataset distillation to imagenet-1k with constant memory.
\newblock In \emph{International Conference on Machine Learning}, pages 6565--6590. PMLR, 2023.

\bibitem[Deng and Russakovsky(2022)]{bptt}
Zhiwei Deng and Olga Russakovsky.
\newblock Remember the past: Distilling datasets into addressable memories for neural networks.
\newblock In \emph{Proceedings of the Advances in Neural Information Processing Systems (NeurIPS)}, 2022.

\bibitem[Dirichlet and Dedekind(1863)]{pigeonhole_principle}
P.G.L. Dirichlet and R. Dedekind.
\newblock \emph{Vorlesungen über Zahlentheorie}.
\newblock English translation: Lectures on Number Theory, American Mathematical Society, 1999 ISBN 0-8218-2017-6, 1863.

\bibitem[Du et~al.(2023)Du, Jiang, Tan, Zhou, and Li]{ftd}
Jiawei Du, Yidi Jiang, Vincent T.~F. Tan, Joey~Tianyi Zhou, and Haizhou Li.
\newblock Minimizing the accumulated trajectory error to improve dataset distillation.
\newblock In \emph{Proceedings of the IEEE/CVF Conference on Computer Vision and Pattern Recognition (CVPR)}, 2023.

\bibitem[Feng et~al.(2023)Feng, Vedantam, and Kempe]{embarrassingly_simple}
Yunzhen Feng, Shanmukha~Ramakrishna Vedantam, and Julia Kempe.
\newblock Embarrassingly simple dataset distillation.
\newblock In \emph{The Twelfth International Conference on Learning Representations}, 2023.

\bibitem[Gidaris and Komodakis(2018)]{convnet}
Spyros Gidaris and Nikos Komodakis.
\newblock Dynamic few-shot visual learning without forgetting.
\newblock In \emph{Proceedings of the IEEE conference on computer vision and pattern recognition}, pages 4367--4375, 2018.

\bibitem[Jubran et~al.(2019)Jubran, Maalouf, and Feldman]{coresets_intro}
Ibrahim Jubran, Alaa Maalouf, and Dan Feldman.
\newblock Introduction to coresets: Accurate coresets.
\newblock \emph{arXiv preprint arXiv:1910.08707}, 2019.

\bibitem[Krizhevsky et~al.(2009)Krizhevsky, Hinton, et~al.]{cifar}
Alex Krizhevsky, Geoffrey Hinton, et~al.
\newblock Learning multiple layers of features from tiny images.
\newblock 2009.

\bibitem[Le and Yang(2015)]{tiny_imagenet}
Ya Le and Xuan Yang.
\newblock Tiny imagenet visual recognition challenge.
\newblock \emph{CS 231N}, 7\penalty0 (7):\penalty0 3, 2015.

\bibitem[Lee et~al.(2022{\natexlab{a}})Lee, Lee, and Hwang]{kfs}
Hae~Beom Lee, Dong~Bok Lee, and Sung~Ju Hwang.
\newblock Dataset condensation with latent space knowledge factorization and sharing.
\newblock \emph{arXiv preprint arXiv:2208.10494}, 2022{\natexlab{a}}.

\bibitem[Lee et~al.(2022{\natexlab{b}})Lee, Chun, Jung, Yun, and Yoon]{dcc}
Saehyung Lee, Sanghyuk Chun, Sangwon Jung, Sangdoo Yun, and Sungroh Yoon.
\newblock Dataset condensation with contrastive signals.
\newblock In \emph{International Conference on Machine Learning}, pages 12352--12364. PMLR, 2022{\natexlab{b}}.

\bibitem[Loo et~al.(2022)Loo, Hasani, Amini, and Rus]{rfad}
Noel Loo, Ramin Hasani, Alexander Amini, and Daniela Rus.
\newblock Efficient dataset distillation using random feature approximation.
\newblock \emph{Advances in Neural Information Processing Systems}, 35:\penalty0 13877--13891, 2022.

\bibitem[Loo et~al.(2023)Loo, Hasani, Lechner, and Rus]{rcig}
Noel Loo, Ramin Hasani, Mathias Lechner, and Daniela Rus.
\newblock Dataset distillation with convexified implicit gradients.
\newblock \emph{arXiv preprint arXiv:2302.06755}, 2023.

\bibitem[Nair and Hinton(2010)]{relu}
Vinod Nair and Geoffrey~E Hinton.
\newblock Rectified linear units improve restricted boltzmann machines.
\newblock In \emph{Proceedings of the 27th international conference on machine learning (ICML-10)}, pages 807--814, 2010.

\bibitem[Nguyen et~al.(2021)Nguyen, Novak, Xiao, and Lee]{kip}
Timothy Nguyen, Roman Novak, Lechao Xiao, and Jaehoon Lee.
\newblock Dataset distillation with infinitely wide convolutional networks.
\newblock \emph{Advances in Neural Information Processing Systems}, 34:\penalty0 5186--5198, 2021.

\bibitem[Radford et~al.(2021)Radford, Kim, Hallacy, Ramesh, Goh, Agarwal, Sastry, Askell, Mishkin, Clark, et~al.]{clip}
Alec Radford, Jong~Wook Kim, Chris Hallacy, Aditya Ramesh, Gabriel Goh, Sandhini Agarwal, Girish Sastry, Amanda Askell, Pamela Mishkin, Jack Clark, et~al.
\newblock Learning transferable visual models from natural language supervision.
\newblock In \emph{International conference on machine learning}, pages 8748--8763. PMLR, 2021.

\bibitem[Rebuffi et~al.(2017)Rebuffi, Kolesnikov, Sperl, and Lampert]{coresets_rebuffi2017}
Sylvestre-Alvise Rebuffi, Alexander Kolesnikov, Georg Sperl, and Christoph~H Lampert.
\newblock icarl: Incremental classifier and representation learning.
\newblock In \emph{Proceedings of the IEEE conference on Computer Vision and Pattern Recognition}, pages 2001--2010, 2017.

\bibitem[Sachdeva and McAuley(2023)]{survey}
Noveen Sachdeva and Julian McAuley.
\newblock Data distillation: A survey.
\newblock \emph{arXiv preprint arXiv:2301.04272}, 2023.

\bibitem[Sucholutsky and Schonlau(2021)]{soft_labels}
Ilia Sucholutsky and Matthias Schonlau.
\newblock Soft-label dataset distillation and text dataset distillation.
\newblock In \emph{2021 International Joint Conference on Neural Networks (IJCNN)}, pages 1--8. IEEE, 2021.

\bibitem[Ulyanov et~al.(2016)Ulyanov, Vedaldi, and Lempitsky]{instance_normalization}
Dmitry Ulyanov, Andrea Vedaldi, and Victor Lempitsky.
\newblock Instance normalization: The missing ingredient for fast stylization.
\newblock \emph{arXiv preprint arXiv:1607.08022}, 2016.

\bibitem[Wang et~al.(2022)Wang, Zhao, Peng, Zhu, Yang, Wang, Huang, Bilen, Wang, and You]{cafe}
Kai Wang, Bo Zhao, Xiangyu Peng, Zheng Zhu, Shuo Yang, Shuo Wang, Guan Huang, Hakan Bilen, Xinchao Wang, and Yang You.
\newblock Cafe: Learning to condense dataset by aligning features.
\newblock In \emph{Proceedings of the IEEE/CVF Conference on Computer Vision and Pattern Recognition}, pages 12196--12205, 2022.

\bibitem[Wang et~al.(2018)Wang, Zhu, Torralba, and Efros]{dataset_distillation}
Tongzhou Wang, Jun-Yan Zhu, Antonio Torralba, and Alexei~A. Efros.
\newblock Dataset distillation.
\newblock \emph{arXiv preprint arXiv:1811.10959}, 2018.

\bibitem[Welinder et~al.(2010)Welinder, Branson, Mita, Wah, Schroff, Belongie, and Perona]{cub200}
Peter Welinder, Steve Branson, Takeshi Mita, Catherine Wah, Florian Schroff, Serge Belongie, and Pietro Perona.
\newblock Caltech-ucsd birds 200.
\newblock 2010.

\bibitem[Zhao and Bilen(2021)]{dsa}
Bo Zhao and Hakan Bilen.
\newblock Dataset condensation with differentiable siamese augmentation.
\newblock In \emph{Proceedings of the International Conference on Machine Learning (ICML)}, pages 12674--12685, 2021.

\bibitem[Zhao and Bilen(2023)]{dm}
Bo Zhao and Hakan Bilen.
\newblock Dataset condensation with distribution matching.
\newblock In \emph{Proceedings of the IEEE/CVF Winter Conference on Applications of Computer Vision (WACV)}, 2023.

\bibitem[Zhao et~al.(2020)Zhao, Mopuri, and Bilen]{dc}
Bo Zhao, Konda~Reddy Mopuri, and Hakan Bilen.
\newblock Dataset condensation with gradient matching.
\newblock \emph{arXiv preprint arXiv:2006.05929}, 2020.

\bibitem[Zhou et~al.(2022)Zhou, Nezhadarya, and Ba]{frepo}
Yongchao Zhou, Ehsan Nezhadarya, and Jimmy Ba.
\newblock Dataset distillation using neural feature regression.
\newblock In \emph{Proceedings of the Advances in Neural Information Processing Systems (NeurIPS)}, 2022.

\end{thebibliography}

\end{document}